\documentclass{article}

    \usepackage[preprint]{neurips_2019}

\usepackage[utf8]{inputenc} 
\usepackage[T1]{fontenc}    
\usepackage{hyperref}       
\usepackage{url}            
\usepackage{booktabs}       
\usepackage{amsfonts}       
\usepackage{nicefrac}       
\usepackage{microtype}      
\usepackage{graphicx}
\usepackage{subfig}
\title{Efficient Object Detection Model for \\
Real-Time UAV Applications}

\author{%
  Subrahmanyam ~Vaddi \\
  Department of Computer Science\\
  Iowa State University\\
  Ames, IA-50014 \\
  \texttt{svaddi@iastate.edu} \\
  \AND
  Ali Jannesari \\
  Department of Computer Science \\
  Iowa State University\\
  Ames, IA-50014 \\
  \texttt{jannesar@iastate.edu} \\
  \And
  Chandan Kumar \\
  Department of Computer Science \\
  Iowa State University\\
  Ames, IA-50014 \\
  \texttt{chandan@iastate.edu} \\
}

\begin{document}

\maketitle

\begin{abstract}
  Unmanned Aerial Vehicles (UAVs) especially drones, equipped with vision techniques have become very popular in recent years, with their extensive use in wide range of applications. Many of these applications require use of computer vision techniques, particularly object detection from the information captured by on-board camera. In this paper, we propose an end to end object detection model running on a UAV platform which is suitable for real-time applications. We propose a deep feature pyramid architecture which makes use of inherent properties of features extracted from Convolutional Networks by capturing more generic features in the images (such as edge, color etc.) along with the minute detailed features specific to the classes contained in our problem. We use VisDrone’18 dataset for our studies which contain different objects such as pedestrians, vehicles, bicycles etc. We provide software and hardware architecture of our platform used in this study. We implemented our model with both ResNet and MobileNet as convolutional bases. Our model combined with modified focal loss function, produced a desirable performance of 30.6 mAP for object detection with an inference time of 14 fps. We compared our results with RetinaNet-ResNet-50 and HAL-RetinaNet and shown that our model combined with MobileNet as backend feature extractor gave the best results in terms of accuracy, speed and memory efficiency and is best suitable for real time object detection with drones.
\end{abstract}

\section{Introduction}

In recent years, autonomous Unmanned Aerial Vehicles (UAVs) especially drones equipped with cameras have become very popular with the wide variety of their applications such as surveillance [16], aerial mapping [17], search and rescue [21], infrastructural inspection [11], precision agriculture [5] etc. Automatic understanding of visual data collected from these drones has been the area of increasing interest. Visual object detection is one of the important aspects of applications of drones and is critical to have in fully autonomous systems. However, the task of object detection with drones is very challenging and complex as it is affected by various imaging conditions such as noise in the images, blur, low resolution, small target sizes etc. The task is even more difficult because of the limited computational resources available on the drones and the need for almost real time performance in many applications such as navigation, traffic management etc. 

Many	UAV studies have been made to detect certain application specific single objects such as vehicles [13], pedestrians [9], autonomous navigation and landing [4] and some studies which detect multiple objects such as [2], [7]. The key challenges in deploying the vision and intelligence capabilities on a UAV platform are 1) to consume minimal power in order to minimize its effect on battery consumption and flight time of the drone 2) require less memory footprint and computational power as typical UAVs have resource limitations 3) process the input data from its camera with low latency and perform faster in order to make critical decisions such as object detection, avoidance and navigation in real time. Conventionally onboard computing infrastructure consists mainly of general purpose machines, such as multicore CPUs and low power microcontrollers. Later the high computational requirement of modern computer vision techniques such as deep learning, led to introduction of massively parallel architectures with prominence of GPUs as accelerators. Even though there are many alternatives to on-board processing such as cloud-centric solutions, they pose additional overheads such as latency in video feed and information security risk. Hence, in this paper we are concerned with developing algorithms based on deep learning for object detection running on UAV equipped with embedded hardware platforms suitable for real time applications.

Traditionally hand-tuned features were used for object recognition and detection. With the breakthrough of deep learning using Convolutional Neural Networks there was a striking performance increase in dealing with these computer vision tasks. The key idea is to learn object model and features from raw pixel data of image. Training these deep learning models typically requires large datasets but this problem has also been overcame by many large scale labeled datasets like ImageNet. Unfortunately, these techniques require huge amount of computational power and built-in memory. There has always been a tradeoff between the detection performances with the speed. Using these techniques on low-cost drones has thus became challenging because of the size, weight and power requirements of these devices. In this paper, we propose an object detection model which is computationally less expensive, memory efficient and fast without compromising the detection performance, running on a drone. We report on experiments measuring mean Average Precision (mAP) and inference time using low cost quadcopter drone as a hardware platform, in the scenario of detecting target objects in an environment containing objects like pedestrians, buses, bicycles etc. We propose a novel Deep Feature Pyramid Network (DFPN) architecture and a modified loss function to avoid class imbalance and achieved real time object detection performance on real drone environment.

\section{Related Work}
\subsection{UAV based Object Detection}

Vision assisted UAV could achieve navigation, path planning and autonomous control for itself. In early studies of UAVs, Saripalli et al. [24] explored the vision-based strategies for autonomous control during UAV landing period. Scaramuzza et al. [25] incorporated vision techniques into their UAV design to enhance the navigation performance. UAVs equipped with computer vision could also offer video or image surveillance services in different application scenarios. Earlier, Ryan et al. [22] in their work gave a general overview towards the implementation of object detection involved computer vision techniques for aerial surveillance. Later many studies focused on data driven applications of UAVs. For example, UAVs been used for smart urban surveillance in [6]. Recently an end to end IoT system based on drones has been proposed in [5] for data driven precision agriculture application. Many recent studies make use of flight control systems utilizing deep learning frameworks for vision assisted tasks to perform scene analysis of aerial images. PIXHAWK [18] is one of the popular flight control systems designed with computer vision algorithms to execute the obstacle detection and autopilot mode during UAV operation, which we have used onboard for our experiments.

\subsection{Object Detection Models}

The first deep learning object detector model was called Overleaf Network [22] which used Convolutional Neural Networks (CNNs) along with the sliding window approach classifying each part of image as object or non-object and combined the results for predictions. This method of using CNNs to solve detection led to new networks being introduced which pushed the state of the art even further. In recent years, numerous object detectors have been proposed by the deep learning community, including Faster R-CNN[23], YOLO[14], R-FCN[12], SSD[28] and RetinaNet[27]. The main  goal  of  these  designs is to improve (1)  the  detection  accuracy measure in terms of mAP and (2) the  computational complexity of their models so that they can achieve real-time performance for embedded and mobile platforms[8]. These detection models can be divided into two categories based on their high-level architecture - the single step approach and the two step (region-based) approach. The single step approach achieved faster results and shown higher memory efficiency whereas the two-step approach achieved better accuracy than former. 

\subsubsection{Region Based Detectors}
Region based detectors divide object detection in two steps. The first step generates a set of regions in the image that have a high probability of being an object. The second step then performs the final detection and classification of objects by taking these regions as input. A prominent example of such a detector is Faster R-CNN [23], in which the first stage, called the Region Proposal Network (RPN), employs the feature extractor of a CNN (e.g. VGG-16, ResNet, etc.) to process images and utilizes the output feature maps of a selected intermediate layer in order to predict bounding boxes of class-agnostic objects on an image. In the second stage, the box proposals are used to crop features of the same intermediate feature maps and pass them through a classifier in order to both predict a class and refine a class-specific box for each proposal. With typically hundreds of proposals per image passed separately through the classifier, Faster R-CNN remains computationally heavy and poses a challenge in achieving high performance in embedded platforms and mobile devices. Other examples of region-based detectors are FPN [26] and R-FCN [12].

\subsubsection{Single Step Detectors}

This class of detectors aims to avoid the performance bottlenecks of the 2-step region-based systems. The YOLO [14] framework casts object detection to a regression problem and in contrast to the RPN + classifier design of Faster R-CNN, employs a single CNN for the whole task. YOLO divides the input image into a grid of cells and for each cell outputs predictions for the coordinates of a number of bounding boxes, the confidence level for each box and a probability for each class. Compared to Faster R-CNN, YOLO is designed for real-time execution and by design provides a trade-off that favors high performance over detection accuracy. In addition, the open-source released version of YOLO has been developed using the C and CUDA based Darknet [20] framework, which enables the use of both CPUs and GPUs and is portable across a variety of platforms. SSD [28] is a single-shot detector aims to combine the performance of YOLO with the accuracy of region-based detectors. SSD extends the CNN architecture of YOLO by adding more convolutional layers and allowing the grid of cells for predicting bounding boxes to have a wider range of aspect ratios in order to increase the detection accuracy for objects of multiple scales. Despite the high detection accuracy and performance, the open-source released version of SSD has been developed using the Caffe framework.

RetinaNet is the latest single step object detector which boasts the state of the art results at this point in time by introducing a novel loss function [27]. This model represents the first instance where one step detectors have surpassed two step detectors in accuracy while retaining superior speed. In this work, we focus on single-shot detectors due to their high performance and applicability to mobile and embedded systems. We have used RetinaNet as our baseline detector for comparison. Our model is based on RetinaNet with certain architectural improvements made for aerial imagery applications. We introduce DFPN architecture embedded within object detector model which is robust and scale invariant.

\section{Methodology}

Our object detection model design features an efficient in-network Deep Feature Pyramid Network (DFPN) architecture and use of anchor boxes to predict outputs. It draws a variety of ideas from [26] and [27]. The reason why one step detectors have lagged behind two step detectors in terms of accuracy was an implicit class imbalance problem that was encountered while training. Class imbalance occurs when the types of training examples are not equal in number. Single step detectors suffer from an extreme foreground/background imbalance, with the data heavily biased towards background examples.

Class imbalance occurs because a single step detector densely samples regions from all over the image. This leads to a high majority of regions belonging to the background class. The two step detectors avoid this problem by using an attention mechanism (RPN) which focuses the network to train on a small set of examples. SSD [28] tries to solve this problem by using techniques like a fixed foreground-background ratio of 1:3, online hard example mining [1] or bootstrapping [10]. These techniques are performed in single step implementations to maintain a manageable balance between foreground and background examples. However, they are inefficient as even after applying these techniques, the training data is still dominated by easily classified background examples. To avoid class imbalance, we used a dynamic loss function as in [27] which down weights the loss contributed by easily classified examples. The scaling factor decays to zero when the confidence in predicting a certain class increases. This loss function can automatically down weight the contribution of easy examples during training and rapidly focus the model on hard examples.

\subsection{Object Detector Model}
We use a simple object detector model by combining the best practices gained from previous research studies.

\paragraph{Input:} An aerial image fed as an input to the model.

\paragraph{Targets:}
The network uses the concept of anchors to predict regions. As our DFPN is integrated with the model, the anchor sizes do not need to account for different scales as that is handled by the multiple feature maps. Each level on the feature pyramid uses 9 anchor shapes at each location. The set of three aspect ratios 1 : 1, 1 : 2, 2 : 1 have been augmented by the factors of 1, $2^1/3, 2^2/3$ for a more diverse selection of bounding box shapes. Similar to the RPN, each anchor predicts a class probability out of a set of K object classes (including the background class) and 4 bounding box offsets. Targets for the network are calculated for each anchor as follows. The anchor is assigned the ground truth class if the IOU (Intersection over Union) of the ground truth box and the anchor $\leq$ 0.5. A background class is assigned if the IOU $\leq$ 0.4 and if the $0.4 \leq$ IOU $\leq$ 0.5, the anchor is ignored during training. Box regression targets are computed by calculating the offsets between each anchor and its assigned ground truth box using the same method used by the Faster RCNN. No targets are calculated for the anchors belonging to the background class, as the model is not trained to predict offsets for a background region. 

\paragraph{Architecture:}
The detector uses a single unified network composed of a backbone network and two task specific subnetworks. The first subnetwork predicts the class of the region and the second subnetwork predicts the coordinate offsets. The architecture is similar to an RPN augmented by an FPN except that we build a deep FPN model as explained below.

\begin{figure}
  \centering
  
  \fbox{ \includegraphics[scale=0.45]{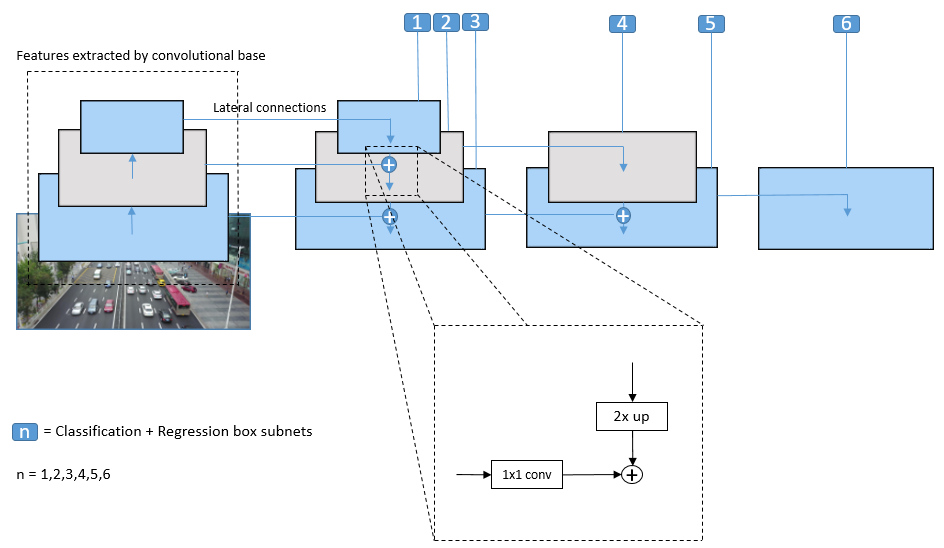}}
  \caption{Proposed Deep Feature Pyramid Network (DFPN) depicting the lateral connections and how the deep pyramids are constructed.}
  \label{fig2}
\end{figure}

In this paper, we introduce a Deep Feature Pyramid Network (DFPN) architecture as shown in Figure~\ref{fig2}. Similar to FPN [26], our goal is to leverage a ConvNet’s pyramidal feature hierarchy, which has semantics from low to high levels, and build deep feature pyramids with high-level semantics throughout. The construction of our pyramid involves a bottom-up pathway, a top-down pathway, and lateral connections. The bottom-up pathway is the feedforward computation of the backbone ConvNet, which computes a feature hierarchy consisting of feature maps at several scales with a scaling step of 2. The topdown pathway hallucinates higher resolution features by up-sampling spatially coarser, but semantically stronger, feature maps from higher pyramid levels. These features are then enhanced with features from the bottom-up pathway via lateral connections. Each lateral connection merges feature maps of the same spatial size from the bottom-up pathway and the top-down pathway. The bottom-up feature map is of lower-level semantics, but its activations are more accurately localized as it was subsampled fewer times. DFPN is built by further creating the FPN structures until the output feature maps converges to 1 creating a deep and rich feature pyramids of image as shown in Figure~\ref{fig2}. We have developed high-level semantic feature maps of all possible scales thus making our solution to be scale invariant. Also, this approach works pretty well with the aerial images, as most of the objects are usually very small in scale. Additionally, our network captures the abrupt changes in scales of objects, which typically occurs in videos captured by drones.

The classification and the regression subnets are quite similar in structure. Each pyramid level is attached with these subnetworks, weights of the heads are shared across all levels. The architecture of the classification subnet consists of a small FCN consisting of 4 convolutional layers of filter size 3 by 3. Each convolutional layer has a relu activation function attached to it and maintains the same channel size as the input feature map. Finally, sigmoid activations are attached to output a feature map of depth A$\star$K where the value for A = 9 and it represents the number of aspect ratios per anchor, K represents the number of object classes. The box regression subnet is identical to the classification subnet except for the last layer. The last layer has the depth of 4$\star$A. The 4 indicates the width, height and x and y coordinate offsets of bounding boxes.

\paragraph{Loss Function:}
The loss of a model is used to train the entire network and it controls the amount by which the gradients are tweaked during back-propagation. A high loss for an object makes the network more sensitive for that object and vice versa. The loss function we used in similar to [27] given by equation~\ref{eq},
\begin{equation}
    F = {\alpha}^2*(1-p)^{\gamma}*c
    \label{eq}
\end{equation}

where c is cross entropy loss, $\alpha$ is the hyper parameter which down weights the loss generated by background class. The value for $\alpha$ could be the inverse class frequency or can be treated as a hyper-parameter to be set during cross validation. We considered square of $\alpha$ as there can be exponential number of background anchors compared to target anchors because of the aerial view of images. $\gamma$ is the exponential scaling factor which down weights the loss contributed by easy examples thereby making the network sensitive only for difficult examples. We considered values of 0.25 for $\alpha$ and 2 for $\gamma$ in our experiments.

\paragraph{Training and Inference:}
Training is performed using Stochastic Gradient Descent (SGD), using an initial learning rate of 0.0001. Horizontal image flipping is the only data augmentation technique used. Inference is performed by running the image through the network. Only the top 1k predictions are taken at each feature level after thresholding the detector confidence at 0.05. Non Maximum Suppression (NMS) is performed using a threshold of 0.5 and the boxes are overlaid on the image to form the final output. This technique is seen to improve training stability for both cross entropy and focal loss in the case of heavy class imbalance. 

\paragraph{Convolutional bases:} Object detectors use convolutional base to create feature maps of images that is embedded with salient information about the image. The accuracy and performance of the detector is highly dependent on how well the convolutional base can capture the meaningful information about the image. The base takes the image through a series of convolutions that make the image smaller and deeper. This process allows the network to make sense of the various shapes in the image. Convolutional bases can be selected on basis of three factors such as speed, accuracy and memory footprint. As per our requirement, we chose ResNet [15] and MobileNet [3] for our studies and compared the results obtained using both. ResNet is one of the popular convolutional bases with the concept of skip connections in convolutional networks. Skip connections add or concatenate features of the previous layer to the current layer. This leads to the network propagating gradients much more effectively during backpropagation. On the other hand, MobileNets are series of convolutional networks best suitable for applications where speed and memory efficiency are of high importance. They are proved to be well suited for embedded system applications. MobileNets introduce depth wise separable convolutions which form their backbone and have led to a speedup in computing the feature map without sacrificing the overall quality.

\section{System Architecture}

We deployed our object detector model on a real drone equipped with onboard GPU and autopilot capable for autonomous operation. In this section we explain the details of hardware and software we used to build our drone.

\subsection{Hardware Setup}

\begin{figure}
  \centering
  
  \fbox{ \includegraphics[scale=0.4]{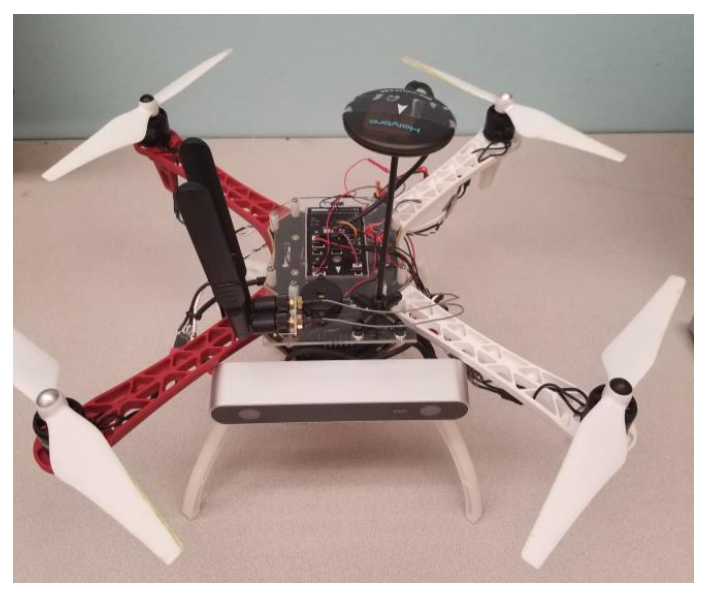}}
  \caption{Jetson TX2 board attached on our DJI Flame Wheel drone.}
  \label{fig1}
\end{figure}

Our drone comprises of DJI Flame Wheel F450 quadcopter with an open source Pixhawk PX4 autopilot mounted on an Orbitty carrier board. We use NVIDIA Jetson TX2 which is a CUDA capable device and runs on Ubuntu Linux4Tegra(L4T) and JetPack-L4T 3.2. It features an integrated 256 CUDA core NVIDIA Pascal GPU, a hex-core ARMv8 65-bit CPU complex, and 8GB of LPDDR4 memory with a 128-bit interface. We have also used a Windows 10, Intel(R) Xeon(R) Silver 4114 CPU @2.20GHz(20 CPUs), 2.2GHz, 32GB RAM with Nvidia Titan Xp GPU for training purposes. Our drone is equipped with ZED stereo camera, which is a standard 3D USB camera for object detection and image capturing. Additionally, it adds depth perception, positional tracking and 3D mapping using advanced sensing technology based on human stereo vision. We use Venom 35C, 14.8V, 5000 mAh lithium polymer battery and Holybro Ublox Neo-M8N GPS module which provides high sensitivity and minimal acquisition time while maintaining low system power. We use a Linux host PC machine to communicate and access the location of drone. Figure~\ref{fig1} shows the hardware setup of our drone.

\subsection{Software Setup}

Jetson TX2 on our drone was first flashed with the latest OS image - Ubuntu 16.04. We further used JetPack installer to install developer tools on the drone. All other libraries useful for computer vision tasks such as TensorRT, cuDNN, CUDA, OpenCV were installed to jumpstart our development environment. We have used ROS Kinetic Kame distribution for hardware abstraction of Joystick and the Controller present on the drone. ROS also plays an important part in controlling devices at a very low-level as well as for transferring data. A ROS system typically consists of a number of independent nodes. For our environment, these nodes are MAVROS node, Controller Node, DNN, Camera and Joystick - each of which is able to communicate with each other using subscribe or publish messaging model. All nodes are registered with the master node (MAVROS in our case) which in turn helps them to find and communicate with each other. The MAVROS enables MAVLink (Micro Air Vehicle Link) 2018 protocol to communicate with the PX4 (Flight Control Unit) on-board.

The on-board Jetson TX2 had Wi-Fi access point enabled by us before installing it on the drone. As a result, the host PC could connect to the Jetson TX2 wirelessly and was able to access it remotely. By sending commands through the terminal, the host was able to control the drone as and when required by Secure Socket Shell (SSH) network protocol. We further use standard UART (Universal Asynchronous Receiver/Transmitter) communications for controls.

\subsection{Dataset}

For our experiments we use VisDrone [19] dataset. VisDrone is the most recent dataset which includes aerial images. Images were captured by different drones flying over 14 different cities separated by thousands of kilometers in China, in different scenarios under various weather and lighting conditions. The dataset consists of 263 video sequences formed by 179,264 frames and 10,209 static images and contains different objects such pedestrian, vehicles, bicycles, etc. and density (sparse and crowded scenes). Frames are manually annotated with more than 2.5 million bounding boxes and some attributes, e.g. scene visibility, object class and occlusion, are provided. VisDrone is available at http://www.aiskyeye.com.

\section{Experimental Results}

In this section, we present a comprehensive quantitative evaluation of our object detection model. The performance of each model is evaluated on the basis of three metrics namely Intersection over Union (IoU), mean Average Precision (mAP) and inference time as frames per second (fps). In the context of object detection IoU metric captures the similarity between the predicted region and the ground truth region for an object present in the image and is calculated as size of intersection of predicted and ground-truth regions divided by their union. Precision is the widely used metric by the object detection community and is defined as the proportion of True Positives among all the detections of system. The mean of average precisions for each category of objects calculated across all the possible IoUs is termed as mAP. Inference time is the rate at which an object detector is capable of processing the incoming camera frames. The rate is calculated as the number of frames per second.

We trained our model on Nvidia Titan Xp GPU machine and deployed it on our onboard Nvidia Jetson TX2 for testing and inference (time taken by the model to predict output of test dataset). Our model was built in two designs. One with ResNet as backbone feature extractor and the other with MobileNet. Table~\ref{table1} shows the comparison of these two designs in terms of their performance, power drawn and network size.

\begin{table}[h!tb] \centering
\caption{Our models performance comparison with ResNet and MobileNet as feature extractors.}
\label{table1}
\begin{tabular}{p{2.8cm}p{2.8cm}ccp{2.8cm}} \hline
Backbones for our model & No. of parameters (in billions) & mAP & Inference time & Inference power draw (in W) \\ \hline
ResNet & 36.7  & 30.6 & 8.6 fps & 120 \\
MobileNet  & 13.5  & 29.2 & 14 fps & 85 \\ \hline
\end{tabular}
\end{table}

From our observations, our model with MobileNet base is very fast in terms of detection inference time compared to ResNet backbone. We achieved a real time speed of 14fps maintaining the quality of detection at 29.2 mAP. Also with MobileNet the overall size of our model decreased by a factor of 3, thereby reducing the memory footprint requirement on drone. We have calculated our power draw values on Nvidia Titan Xp GPU with a maximum possible power draw of 300W. Average power drawn with idle GPU is 15W. Our results show that our model combined with MobileNet as feature extractor draws very less power compared to the other models which results in long battery life of the drone.

\begin{figure}
  \centering
  
  \fbox{\includegraphics[width=6.5cm, height=4.5cm]{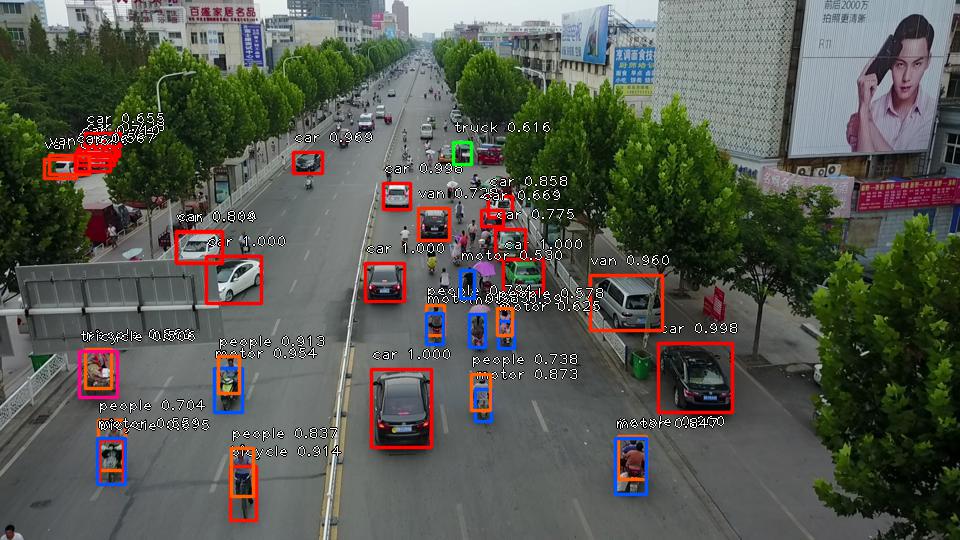}}  \fbox{\includegraphics[width=6.5cm, height=4.5cm]{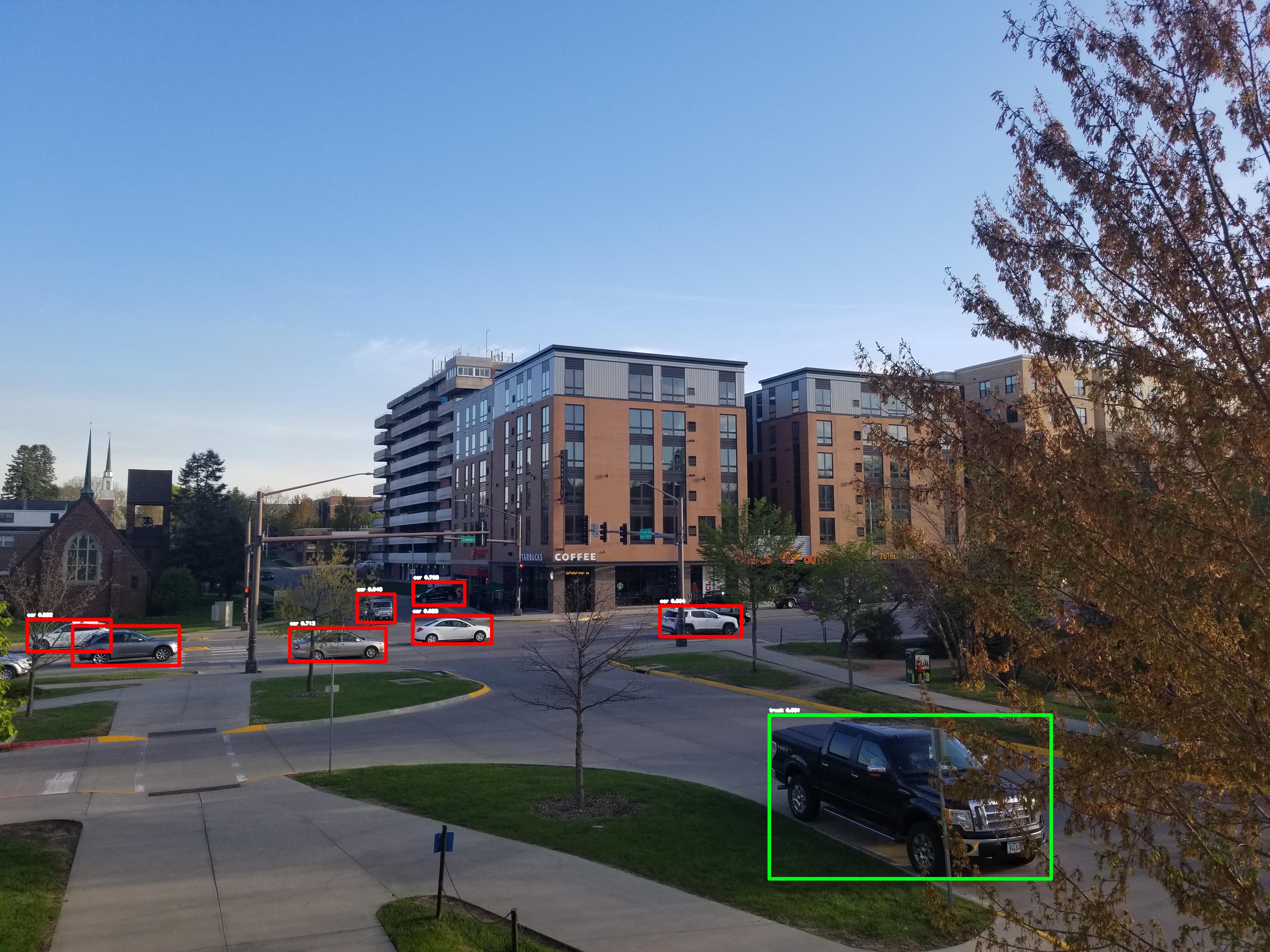} }
  \caption{Our models prediction output on a) one of the aerial images in VisDrone’18 dataset and b) one of the frames captured by our drone.}
  \label{fig3}
\end{figure}

Table~\ref{table2} shows the performance comparison of two models, RetinaNet and HAL-RetinaNet (the top performer of VisDrone 2018 object detection challenge) with our model. The results are evaluated on the basis of mean Average Precision on VisDrone 2018 validation and test data aerial images. From the table 1, we can see that our model operates at 8.6 frames per second which is well suited for real time object detection applications. Our deep feature pyramid network along with the modified loss function gave a better mean average precision of 30.6. Figure~\ref{fig3} shows the object detection result on one of the aerial images in VisDrone dataset and result on one that is captured by our drone.

\begin{table}[h!tb] \centering
\caption{Comparison of object detection performance between RetinaNet, HAL-RetinaNet and our model in terms of mAP and inference time.}
\label{table2}
\begin{tabular}{ccc} \hline
Model & mAP & Inference time \\ \hline
RetinaNet & 23 & 8 fps \\
HAL-RetinaNet   & 31.8  & N/A \\ 
Our Model (With MobileNet) & 29.2 & 14 fps \\ \hline
\end{tabular}
\end{table}

\section{Conclusion and Future Work}

UAVs equipped with intelligence and vision techniques such as object detection have wide variety of real-world applications. Drones require memory efficient, computationally less expensive and fast solutions for the task of object detection. In this paper we proposed a Deep Feature Pyramid Network (DFPN) architecture embedded in object detector model combined with modified focal loss function to avoid problem of inherent class imbalance and improve detection capability even with varying scales of images. We considered two networks namely ResNet and MobileNet as backbone convolutional bases for our detection model. Even though our model combined with ResNet provided better results in terms of detection accuracy, combination of MobileNet resulted in real time speeds without compromising the detection accuracy. We also compared our model with RetinaNet-ResNet50 and HAL-RetinaNet and shown that our model produced overall better results in terms of accuracy, speed and memory efficiency. We deployed our model on a real drone and were successfully able to detect objects in real time.

Potential future works include performance improvements in terms of object detection by applying finer-level optimizations such as reducing creation of redundant anchors, focusing more on missing occlusions etc. Safety is another major concern in terms of drones. There can be many situations such as collision with the objects, external disturbances like winds, chances of drone moving out of manual controller zone, battery issues, chances of getting stolen and other safety hazards. We will be implementing these external drone related safety features in our future work. Further, we focus on making our drone more intelligent by implementing features such as object tracking and path planning.

\section*{References}

\small

[1] A. Shrivastava, A. G. (2016). Training region-based object detectors with online hard example mining In CVPR.

[2] al, T. T. (2017). Fast Vehicle Detection in UAV Images. International Workshop on Remote Sensing with Intelligent Processing (RSIP). 

[3] Andrew G Howard et al. (2017). MobileNets: Efficient Convolutional Neural Networks for Mobile Vision Applications. CoRR. 

[4] C. Forster, M. F. (2015). Continuous on-board monocular-vision-based elevation mapping applied to autonomous landing of micro aerial vehicles. Proc. IEEE International Conference on Robotics and Automation (ICRA),pp 111-118. 

[5] Chandra, R. (2018). FarmBeats: Automating Data Aggregation. Farm Policy Journal Vol 15: pp. 7-16. 

[6] Chen, N. C. (2017). Enabling smart urban surveillance at the edge. In Smart Cloud (SmartCloud). IEEE International Conference on (pp. 109- 119). 

[7] Christos Kyrkou, G. P. (2018). DroNet: Efficient convolutional neural network detector for real-time UAV applications. Design, Automation and Test in Europe Conference and Exhibition (DATE). 

[8] G. D. T. N. Subarna Tripathi, B. K. (2017). Low-complexity object detection with deep convolutional neural network for embedded systems. Proc.SPIE, vol. 10396, pp. 10 396 – 10 396 – 15.

[9] H. Lim, a. S. (2015). Monocular Localization of a moving person onboard a Quadrotor MAV. Proc. IEEE International Conference on Robotics and Automation (ICRA), pp. 2182-2189.

[10] H. Rowley, S. B. (1995). Human face detection in visual scenes. Technical Report CMU-CS-95-158R, Carnegie Mellon University.

[11] I. Sa, S. H. (2015). Outdoor flight testing of a pole inspection UAV incorporating high-speed vision. Springer Tracts in Advanced Robotics, pp. 107-121. 

[12] J. Dai, Y. L. (2016). R-FCN: Object Detection via Regionbased Fully Convolutional Networks. NIPS, pp. 379–387. 

[13] J. Gleason, A. N. (2011). Vehicle detection from aerial imagery. Proc. IEEE International Conference on Robotics and Automation (ICRA), pp. 2065-2070. 

[14] J. Redmon, S. D. (2016). You Only Look Once: Unified, Real-Time Object Detection. IEEE Conference on Computer Vision and Pattern Recognition (CVPR), pp. 779–788. 

[15] K.He, X. J. (2016). Deep residual learning for image recognition. CVPR. 

[16] M. Bhaskaranand, a. J. (n.d.). Low-complexity video encoding for UAV reconnaissance and surveillance. Proc. IEEE Military Communications Conference (MILCOM), pp. 1633-1638, 2011. 

[17] Madawalagama, S. (2016). Low Cost Aerial Mapping with Consumer Grade Drones. 37th Asian Conference on Remote Sensing. 

[18] Meier, L. T. (2012). PIXHAWK: A micro aerial vehicle design for autonomous flight using onboard computer vision. Autonomous Robots, 33(1-2), pp.21-39. 

[19] Pengfei Zhu, L. W. (2018). Vision meets drones: A challenge. CoRR, abs/1804.07437. URL http://arxiv.org/abs/1804.07437. 

[20] Redmon, J. (2016). Darknet: Open Source Neural Networks in C. http: //pjreddie.com/darknet/. 

[21] Rudol, P. D. (2007). A UAV search and rescue scenario with human body detection and geolocalization. Advances in Artificial Intelligence, pp. 1-13. 

[22] Ryan, A. Z. (2004). An overview of emerging results in cooperative UAV control. 43rd IEEE Conference on (Vol. 1, pp. 602-607).

[23] S. Ren, K. H. (2017). Faster R-CNN: Towards Real-Time Object Detection with Region Proposal Networks. IEEE Transactions on Pattern Analysis and Machine Intelligence, vol. 39, no. 6, pp. 1137–1149. 

[24] Saripalli, S. M. (2002). Vision-based autonomous landing of an unmanned aerial vehicle. Robotics and automation,Proceedings. ICRA'02. IEEE international conference on (Vol. 3, pp. 2799-2804). IEEE. 

[25] Scaramuzza, D. A. (2014). Vision-controlled micro flying robots: from system design to autonomous navigation and mapping in GPS-denied environments. IEEE Robotics and Automation Magazine. 

[26] T.-Y. Lin, P. D. (2017). Feature pyramid networks for object detection. CVPR. 

[27] Tsung-Yi Lin, P. G. (2017). Focal Loss for Dense Object Detection. ICCV. 

[28] W. Liu, D. A.-Y. (2016). SSD: Single Shot MultiBox Detector. ECCV, pp. 21–37.

\end{document}